\documentclass[10pt,journal]{IEEEtran}

\ifCLASSOPTIONcompsoc
  \usepackage[nocompress]{cite}
\else
  \usepackage{cite}
\fi
\usepackage[table]{xcolor}
\usepackage{graphicx}
\usepackage{amssymb}
\usepackage{amsmath}
\usepackage{multirow}
\usepackage{tabularx}
\usepackage{dsfont}
\usepackage{url}
\usepackage{color}
\usepackage{amsthm}
\usepackage[ruled,linesnumbered]{algorithm2e}
\usepackage[export]{adjustbox}
\usepackage{booktabs}
\usepackage{subcaption}
\usepackage[font=footnotesize,labelfont=bf]{caption}
\usepackage{longtable}
\usepackage{tabu}
\usepackage{diagbox,booktabs}
\usepackage{multicol}

\def\argmin{\operatornamewithlimits{arg\,min}}
\def\argmax{\operatornamewithlimits{arg\,max}}

\usepackage[numbers]{natbib}

\definecolor{darkgreen}{rgb}{0,0.694,0.298}
\definecolor{purple}{rgb}{0.4,0.176,0.569}
\definecolor{royalblue}{RGB}{65,105,225}
\definecolor{americanrose}{rgb}{1.0, 0.01, 0.24}
\definecolor{applegreen}{rgb}{0.55, 0.71, 0.0}

\newcommand{\figref}[1]{Fig.~\ref{#1}}
\newcommand{\reqref}[1]{Eq.~\eqref{#1}}
\newcommand{\secref}[1]{Sec.~\ref{#1}}
\newcommand{\tableref}[1]{Table~\ref{#1}}

\makeatletter
\DeclareRobustCommand\onedot{\futurelet\@let@token\@onedot}
\def\@onedot{\ifx\@let@token.\else.\null\fi\xspace}
\def\eg{\emph{e.g}\onedot} 
\def\ie{\emph{i.e}\onedot} 
 
\def\etc{\emph{etc}\onedot}

\makeatother

\definecolor{americanrose}{rgb}{1.0, 0.01, 0.24}

\definecolor{citecolor}{RGB}{65,105,225}
\usepackage[breaklinks=true,colorlinks,citecolor=citecolor,bookmarks=false]{hyperref}

\usepackage{pifont}

\begin{document}
%
\title{Adversarial Relighting Against Face Recognition}
%
%
%
%

\author{Qian~Zhang,~\IEEEmembership{Student Member,~IEEE,}
        Qing~Guo,~\IEEEmembership{Member,~IEEE,}
        Ruijun~Gao,~\IEEEmembership{Student Member,~IEEE,}
        Felix~Juefei-Xu,~\IEEEmembership{Member,~IEEE,}
        Hongkai~Yu,~\IEEEmembership{Member,~IEEE,} 
        Wei~Feng$^\dagger$,~\IEEEmembership{Member,~IEEE}
\IEEEcompsocitemizethanks{\IEEEcompsocthanksitem Qian Zhang, Qing Guo, Ruijun Gao, and Wei Feng$^\dag$~(corresponding author, E-mail: wfeng@ieee.org) are with the School of Computer Science and Technology, College of Intelligence and Computing, Tianjin Key Laboratory of Cognitive Computing and Application, Tianjin University, Tianjin 300305, China, and are also with the Key Research Center for Surface Monitoring and Analysis of Cultural Relics (SMARC), State Administration of Cultural Heritage, China. 
Qing Guo is also with the Nanyang Technological University, Singapore.
Felix Juefei-Xu is with Alibaba Group, USA.
Hongkai Yu is with Cleveland State University, Cleveland, OH, USA. 
}

}

%
%

\markboth{IEEE Transcations on Image Processing, In Submission}%
{Shell \MakeLowercase{\textit{et al.}}: Bare Demo of IEEEtran.cls for Computer Society Journals}
%


\IEEEtitleabstractindextext{%
\begin{abstract}
Deep face recognition (FR) has achieved significantly high accuracy on several challenging datasets and fosters successful real-world applications, even showing high robustness to the illumination variation that is usually regarded as a main threat to the FR system.
However, in the real world, illumination variation caused by diverse lighting conditions cannot be fully covered by the limited face dataset.
In this paper, we study the threat of lighting against FR from a new angle, \ie, \textit{adversarial attack}, and identify a new task, \ie, \textit{adversarial relighting}. Given a face image, adversarial relighting aims to produce a naturally relighted counterpart while fooling the state-of-the-art deep FR methods.
To this end, we first propose the physical model-based adversarial relighting attack (ARA) denoted as \textit{albedo-quotient-based adversarial relighting attack (AQ-ARA)}. It generates natural adversarial light under the physical lighting model and guidance of FR systems and synthesizes adversarially relighted face images.
Moreover, we propose the \textit{auto-predictive adversarial relighting attack (AP-ARA)} by training an adversarial relighting network (ARNet) to automatically predict the adversarial light in a one-step manner according to different input faces, allowing efficiency-sensitive applications.
More importantly, we propose to transfer the above digital attacks to \textit{physical ARA (Phy-ARA)} through a precise relighting device, making the estimated adversarial lighting condition reproducible in the real world.
We validate our methods on three state-of-the-art deep FR methods, \ie, FaceNet, ArcFace, and CosFace, on two public datasets.
The extensive and insightful results demonstrate our work can generate realistic adversarial relighted face images fooling face recognition tasks easily, revealing the threat of specific light directions and strengths.
\end{abstract}

}

\maketitle
\IEEEdisplaynontitleabstractindextext

%
\IEEEpeerreviewmaketitle

\section{Introduction}
\label{sec:intro}

The fast-paced development of deep learning (DL) has bolstered the deployment of high-performance DL-based face recognition systems (FRS) \cite{ding2015robust,hu2020disentangled}. Compared to non DL-based FRS from a decade ago, the DL-based FRS nowadays can handle more challenging unconstrained scenarios and is very well suited for handling various FR tasks in unconstrained real-world scenarios, especially when faces are under various known or unknown degradations such as being very low-resolution \cite{zou2011very,abiantun2019ssr2,juefei2016deepgender}, at an off-angle pose \cite{yi2013towards,juefei2015spartans,pal2016discriminative}, heavily occluded by objects or crowd \cite{huang2021face,juefei2014hallucinating,juefei2016fastfood}, \etc. Among the mentioned degradation factors, illumination variations is perhaps one of the most challenging one due to its high variability and pervasiveness for faces across an array of ConOps scenarios due to the collective effects from complex environmental lighting as well as the facial structure and reflectance properties.

In order to mitigate the challenges that illumination variations have posed on FRS, face relighting was proposed to adjust the lighting appearance on a given face image. More specifically, face relighting aims at altering how light and shadow are cast on the face based on a desired illumination that is usually a result of diffuse or directional lighting. Traditionally, face relighting requires decomposition of the face image into shape geometry, lighting, and reflectance maps, respectively, and then novel lighting, \ie, relighting, is achieved by swapping the intrinsic lighting map with the desired one. However, the accuracy of such relighting and its realisticity depends on the precise estimation of the geometry and reflectance map, which is a difficult task to accomplish, especially when the faces are not under ideal studio setting, but instead in an unconstrained environment. DL-based face relighting methods attempt to execute the aforementioned process in an end-to-end or semi-end-to-end fashion by leveraging large-scale training image pairs. In these recent studies \cite{sun2019single,wang2020single,stoschek2000image,zhang2021tmm}, we have seen a jump in face relighting performance both quantitatively and qualitatively, compared to traditional non-DL-based relighting methods.

%
\begin{figure}[t]
\centering
\includegraphics[width=\linewidth]{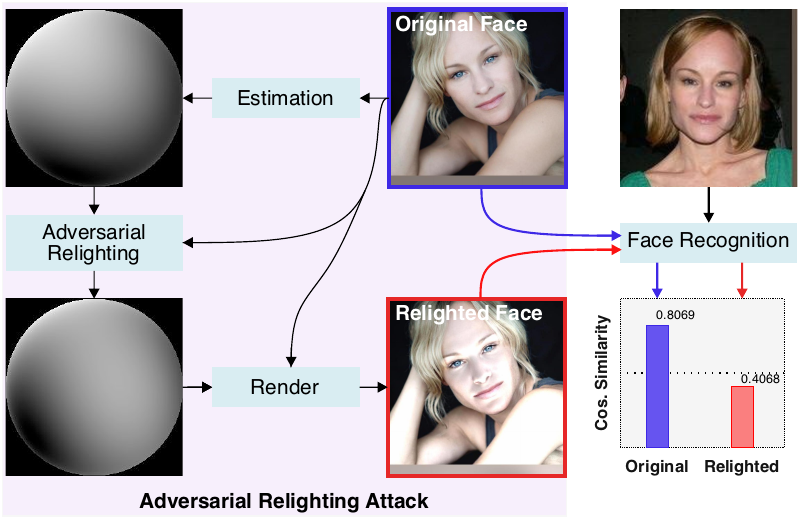}
\caption{Intuitive idea of the new task \textit{adversarial relighting attack (ARA)}. The original face is relighted via the ARA, making the face recognition system fail to identify the same person, that is, the cosine similarity reduces from 0.8069 to 0.4068.}
\label{fig:motivation}
\end{figure}

By capitalizing the advances in DL-based face relighting capabilities, in this work, we are proposing a new study that aims at revealing the vulnerabilities of FRS from the angle of face relighting. To be more specific, we have identified a new task, \ie, adversarial face relighting attack, whose goal is, given a source face image, to produce a naturally relighted counterpart while fooling the state-of-the-art deep FR methods (See \figref{fig:motivation} for the intuitive idea).
\textit{\bf First}, we propose the physical model-based adversarial relighting attack (ARA), that is, \textit{albedo-quotient-based adversarial relighting attack (AQ-ARA)}. Specifically, we define an adversarial objective function based on the physical lighting model and tune the lighting parameters by maximizing the function. 
As a result, we can generate natural adversarial light under the said physical lighting model and the guidance of FRS and synthesize the adversarially relighted face images.
\textit{\bf Second}, we design the auto-predictive adversarial relighting attack (AP-ARA) by training an adversarial relighting network (ARNet) to automatically predict the adversarial light in a one-step manner according to different input faces, allowing efficiency-sensitive applications. 
\textit{\bf Third}, we propose to transfer the aforementioned digital adversarial attacks to \textit{physical ARA (Phy-ARA)} through a precise relighting device, making the estimated adversarial lighting condition reproducible in the real world.
We validate our methods on three state-of-the-art deep FR methods, \ie, FaceNet \cite{schroff2015facenet}, ArcFace \cite{deng2019arcface}, and CosFace \cite{wang2018cosface}, on two public face recognition datasets.
\textit{\bf More importantly}, we conduct both digital and physical experiments to analyze the effects of light to FR via our ARAs, revealing and validating the threat of challenging lighting conditions.
To the best of our knowledge, this work is the very first attempt to study how face relighting can be capitalized to adversarially affect the FRS from the angle of the proposed adversarial relighting attacks.

\section{Related Work}\label{sec:related}

\subsection{Relighting methods.}
In the areas of computer vision and graphics, relighting is an effective way to adjust the illumination variations for an enhanced or different-style visualization. For example, face relighting could adjust the lighting appearance on one face image to generate new portrait images \cite{zhou2019deep}. \cite{shashua2001quotient} re-renders the input front-view image to simulate new illumination conditions given a sample of reference images with varying illumination conditions. Based on a combination of low-level image features and image morphing, \cite{stoschek2000image} is able to synthesize the re-illuminated face images in different pose angles. \cite{wang2020single} generates the photorealistic portrait relighting image based on a reference image or an environment map. \cite{sun2019single} produces a portrait relighting image with an encoder-decoder based CNN network where the target illumination is incorporated in the bottleneck of the network. \cite{zhang2021tmm} provides a reinforcement learning-based approach to portrait relighting. Different with the existing methods, this paper is to design a new deep adversarial relighting method to attack the face recognition.   

\subsection{Attacks against face recognition.}
The attacks against face recognition might make the current face recognition systems  vulnerable. They could be roughly divided into four classes \cite{singh2020robustness}: Perturbing, Morphing, Retouching, Tampering. Perturbing attack adds imperceptible perturbations so as to fool the face recognition system. \cite{goswami2018unravelling} shows that several deep learning based commercial face recognition algorithms and systems are vulnerable to different adversarial perturbation attacks. Morphing attack is to generate a morphed face with the  imperceptible adversarial attack embedded for face recognition. \cite{ferrara2014magic,agarwal2017swapped} shows that some of current commercial face recognition systems cannot protect users from morphed faces. The facial retouching attack is also possible to fool some commercial face recognition systems~\cite{bharati2016detecting}. Recently, generator adversarial networks (GANs) based fake image synthesis and the  well-known DeepFake method~\cite{li2018exposing,tolosana2020deepfakes} could generate or modify the face images, leading to the tampering based attack to face recognition. Different with the above methods, the proposed adversarial relighting method attacks the face recognition from a new perspective of face relighting. 

\subsection{General adversarial attacks.}

Recently, the adversarial attack to fool deep neural networks has attracted many research attentions. Given an image as input, the adversarial attack can be realized by adding imperceptible noises or applying natural transformations. On the one hand, several adversarial noise attack methods got promising attack results, such as  gradient computation based fast gradient sign method (FGSM) \cite{fgsm}, iterative-version FGSM \cite{ifgsm}, momentum iterative FGSM \cite{mifgsm}, different distance metrics based C\&W method \cite{cw}, attended regions and features based TAA method \cite{gao2021tmm}, randomization based \cite{san2020tmm}, perceptually aware and stealthy adversarial denoise \cite{cheng2021tmm}, and so on. On the other hand, some natural transformations that are imperceptible to humans can be applied for image attack. For example, the adversarial attacks can be implemented with various transformations, \eg, semantic-aware colorization or texture transfer \cite{cadv}, motion blurring synthesis \cite{abba}, watermark overlap \cite{advwatermark}, rain \cite{advrain} and haze synthesis \cite{gao2021advhaze}, \etc. By adding the adversarial relighting for attack, the adversarial relighting method proposed in this paper is a kind of novel natural transformation based adversarial attack method.

\section{Adversarial Relighting Attack}\label{sec:method}
In this section, we propose two adversarial relighting attacking (ARA) methods. The first one is based on the physical model and denoted as the albedo-quotient-based ARA (AQ-ARA) (see \secref{subsec:aq-ara}). The second one uses the CNN to automatically predict the adversarial light in an one-step way (\ie, auto-predictive-based ARA (AP-ARA)) (see \secref{subsec:ap-ara}). With these two methods, we further design a physical ARA \secref{subsec:phy-ara} to reproduce the adversarial light in the real world.

\subsection{Albedo-Quotient-based ARA (AQ-ARA)} \label{subsec:aq-ara}
\begin{figure}[t]
\centering
\includegraphics[width=\linewidth]{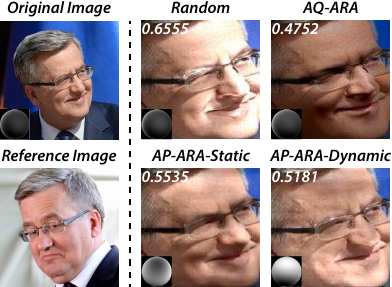}
\caption{An relighting example of Random, AQ-ARA, and AP-ARA (both static and dynamic). The cosine similarity between each relighted face and the reference face based on FaceNet are showed at the left-hand corner.}
\label{fig:example}
\end{figure}

Given a face image $\mathbf{I}$, we assume it follows the Lambertian model that is a widely used face rendering model. Thus, we can represent the face image as
%
\begin{align}\label{eq:lamb}
\mathbf{I} = \mathbf{R}\odot \text{f}(\mathbf{N},\mathbf{L}),
\end{align}
%
where $\mathbf{R}$, $\mathbf{N}$, and $\mathbf{L}$ denote the reflectance, normal, and lighting, respectively. The function $\text{f}(\cdot)$ is the Lambertian shading function. 
More specifically, $\mathbf{L}$ is a nine-dimensional vector corresponding to the nine spherical harmonics coefficients. 
Our objective is to update the lighting (\ie, $\mathbf{L}$) to a new one (\ie, $\hat{\mathbf{L}}$) and produce a new face $\hat{\mathbf{I}}$ that can mislead a state-of-the-art face recognition (FR) method. We denote $\hat{\mathbf{L}}$ as the \textit{adversarial lighting}. This new task actually combine the face relighting with the adversarial attack, thus we name it as the \textit{adversarial relighting attack}. According to \reqref{eq:lamb}, we need to estimate reflectance and normal, which is not easy since calculating the accurate reflectance map is still an open problem. 
To alleviate the requirement, we borrow the albedo-quotient image introduced in \cite{shashua2001quotienta} to get the reflectance-free method for relighting.   
Specifically, we can represent the same face with different lighting conditions as $\mathbf{I} = \mathbf{R}\odot \text{f}(\mathbf{N},\mathbf{L})$ and $\hat{\mathbf{I}} = \mathbf{R}\odot \text{f}(\mathbf{N},\hat{\mathbf{L}})$. As a result, we have
%
\begin{align}\label{eq:quatient}
\hat{\mathbf{I}} =  \mathbf{R}\odot \text{f}(\mathbf{N},\hat{\mathbf{L}})
             =  \frac{\text{f}(\mathbf{N},\hat{\mathbf{L}})}{ \text{f}(\mathbf{N},\mathbf{L})}\mathbf{I}.
\end{align}
%
With \reqref{eq:quatient}, we can relight the face image $\mathbf{I}$ through the normal $\mathbf{N}$, the original lighting $\mathbf{L}$, and the targeted light $\hat{\mathbf{L}}$ \cite{zhou2019deep}.
%
We first estimate face normal $\mathbf{N}$ and lighting $\mathbf{L}$ following the implementation of \cite{zhu2017face,zhou2019deep}. Then we define the objective function to estimate the adversarial lighting. 
Given a deep neural network $\phi(\cdot)$ for the FR, we aim to calculate the adversarial lighting $\hat{\mathbf{L}}$ to let the relighted face image $\hat{\mathbf{I}}$ mislead the FR by solving
%
\begin{align}\label{eq:aqara}
\hat{\mathbf{L}} = & \argmin_{\mathbf{L}'}{\text{sim}(\phi(\frac{\text{f}(\mathbf{N},\mathbf{L}')}{ \text{f}(\mathbf{N},\mathbf{L})}\mathbf{I}),\phi(\mathbf{I}))}, \nonumber \\
& \text{subject to}~\|\mathbf{L}'-\mathbf{L}\|_{\infty}\leq \epsilon
\end{align}
%
where $\text{sim}(\cdot)$ denotes the cosine similarity function and $\phi(\cdot)$ is the embedding of the input face images. The $\epsilon$ controls the changing degrees of lighting.
Intuitively, we minimize the similarity by tuning the lighting, that is, to let the relighted face be different from the original counterpart under the constraint of $\epsilon$. 
We can calculate the gradient of the loss function with respect to light to realize the gradient-based attack. As a result, the attack method can be integrated into any gradient-based additive-perturbation attack methods, \eg, FGSM, BIM, MIFGSM.
Here, we use the sign gradient descent optimization with the step size $\lambda=\frac{\epsilon}{T}$. $T$ denotes the iteration number and we fix it as ten as a common setup in adversarial attacks. 
We show an example of AQ-ARA in \figref{fig:example}. Compared with the random relighting, the adversarial relighting lets the similarity with the reference image decrease significantly while having realistic appearance. The random relighting means we uniformly sample a light $\mathbf{L}$ within the range of $[-\epsilon,\epsilon]$ and relight the face via \reqref{eq:quatient}.

%
%

\subsection{Auto-Predictive-based ARA (AP-ARA)}\label{subsec:ap-ara}

Although above method is able to achieve effective adversarial relighting, the iterative optimization manner limits the potential applications in particular for efficiency-sensitive applications. In the following, we propose an adversarial relighting network (ARNet) to adaptively predict the adversarial lighting in an one-step way.

Given the original lighting $\mathbf{L}$, we use a deep neural network to map the $\mathbf{L}$ to the adversarial lighting $\hat{\mathbf{L}}$, directly. To realize this goal, a naive solution is to build the lighting pair dataset through the method introduced in \secref{subsec:aq-ara}. Nevertheless, this strategy might cost a lot of time due to the iteration operation. 
To alleviate this problem, we propose an end-to-end network  denoted as adversarial relighting network (ARNet) containing three stages, \eg, predicting the original light, estimating the adversarial lighting based on the original one, and relighting the face under the adversarial lighting. We can train this network under the supervision of cosine similarity function directly and do not require the lighting pair dataset. 

%
\begin{figure*}[t]
\centering
\includegraphics[width=\linewidth]{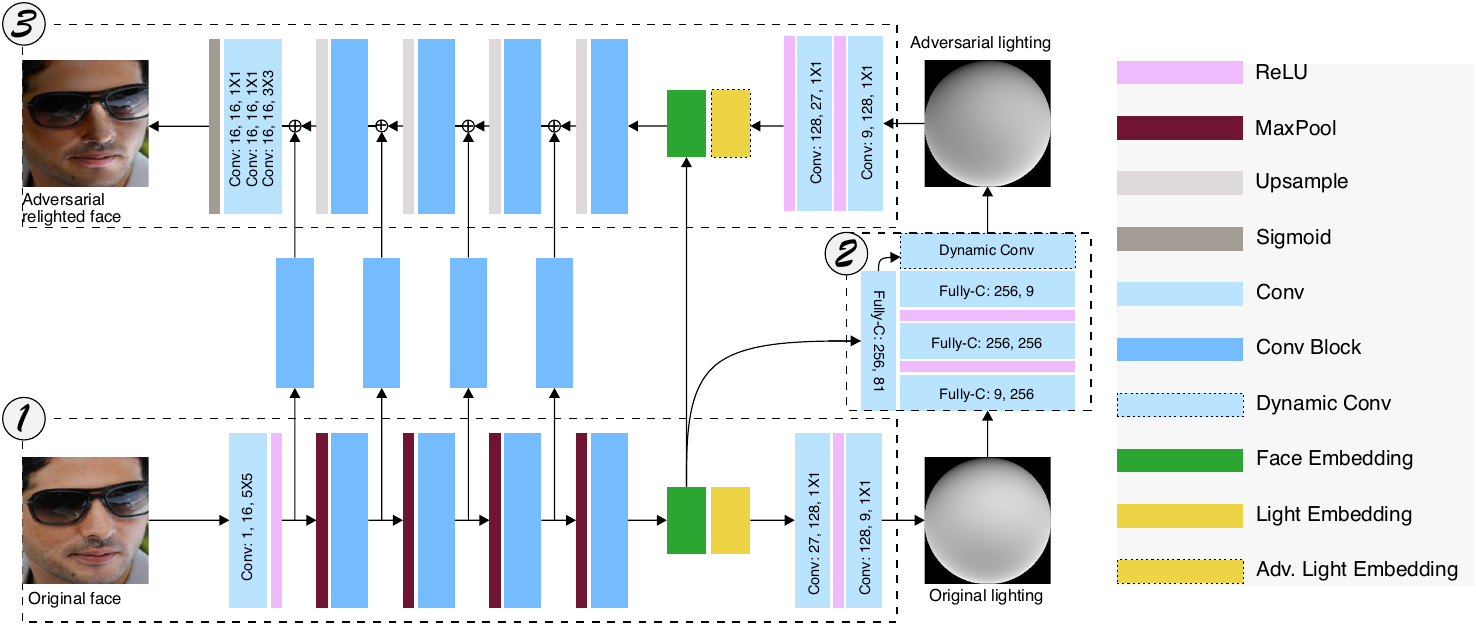}
\caption{Architecture of the proposed adversarial relighting network (ARNet). It contains three modules, \ie, \ding{172} lighting prediction network (LPreNet), \ding{173} adversarial lighting estimation network (AdvLNet), and \ding{174} lighting rendering network (LRenNet). The `Conv Block' contains two convolution layers with the size of $3\times 3$. The first convolutional layer is followed by a BatchNorm layer and a ReLU layer while the second one is only followed by a BatchNorm layer. The `Dynamic Conv' means that the weights of the convolution layer is estimated from a fully connection layer, which makes the adversarial lighting adapt to different face embeddings.}
\label{fig:pipeline}
\end{figure*}
%

{\bf Adversarial relighting network (ARNet).} 
The proposed ARNet contains three modules, \ie, \textit{lighting prediction network (LPreNet)}, \textit{adversarial lighting estimation network (AdvLNet)}, and \textit{lighting rendering network (LRenNet)}. Given a face image $\mathbf{I}$, we first feed it to the LPreNet and get the original lighting parameters $\mathbf{L}$, the face embedding and light embedding, respectively, as the stage \ding{172} shown in \figref{fig:pipeline}.
Then, we use AdvLNet to estimate the adversarial lighting based on the three outputs (See the stage \ding{173} in \figref{fig:pipeline}). Finally, we render the relighted face via the LRenNet by concatenating the embeddings of face and adversarial lighting as inputs, as the stage \ding{174} shown in \figref{fig:pipeline}. 
For the LPreNet and LRenNet, we regard it as the hourglass network \cite{zhou2019deep}. The AdvLNet serves as a transformation for the light embedding, mapping the original light to the adversarial counterpart.
We construct the AdvLNet with three fully connection layers that are linked by two relu layers.
Moreover, to let AdvLNet adapt to different face appearances, we propose to add a dynamic convolution layer whose convolutional weights are estimated by a fully connection layer and the face embedding.
%
%

{\bf Loss functions and training details.} 
We train the three networks in a two-stage way. For the first stage, we train the LPreNet and LRenNet by regarding them as a pure deep relighting task, excluding the AdvLNet. Following the setups of \cite{zhou2019deep}, we use the CelebA-HQ dataset \cite{karras2018progressive} and generate a training example by randomly selecting an original image $\mathbf{I}$ and a targeted relighting image $\mathbf{I}_t$ with their corresponding ground truth lighting $\mathbf{L}$ and $\mathbf{L}_t$, respectively. The LPreNet is fed with the original image $\mathbf{I}$ and predicts the lighting parameters $\mathbf{L}^*$; the LRenNet takes the targeted lighting parameters $\mathbf{L}_t$ and the face embedding of $\mathbf{I}$ as inputs and estimate the relighting counterpart $\mathbf{I}_t^*$.
Under the supervision of the ground truths of relighted image and light, \ie, $\mathbf{I}_t$ and $\mathbf{L}_t$, we can train LPreNet and LRenNet in an end-to-end strategy.
Please refer to \cite{zhou2019deep} for details.
%

After the first stage, we get the LPreNet and LRenNet for lighting prediction and rendering, respectively. For the second stage, we fix the two pre-trained networks, and tune the parameters of AdvLNet under the supervision of a FR method, \eg, $\phi(\cdot)$, with the following loss function
%
\begin{align}\label{eq:lossadv}
\mathcal{L}_\text{adv} = \text{sim}(\phi(\hat{\mathbf{I}}),\phi(\mathbf{I})) + \frac{1}{N}\|\hat{\mathbf{I}}-\mathbf{I}\|_1,
\end{align}
%
where $\hat{\mathbf{I}}=\text{LRenNet}(\text{AdvLNet}(\text{LPreNet}(\mathbf{I})))$, and $N$ is the number of pixels. The first term is the same with the \reqref{eq:aqara} and make sure the relighted face can fool the FR method. The second term is to limit the potential face variation after adversarial relighting. In the field of adversarial attack, it is desired to have small variation on the original image while misleading the targeted FR method.  

In terms the training details, we use the datasets of VGGFace2 \cite{cao2018vggface2} or CelebA \cite{liu2015deep} and employ the stochastic gradient descent (SGD) optimizer with learning rate ${10}^{-3}$ and momentum $0.9$ to update AdvNet's parameters. We use mini-batch training strategy with batch size $8$ and train AdvLNet with ten epochs. Please refer to \secref{subsec:setups} for more details on datasets for training.

\subsection{Physical ARA (Phy-ARA)} \label{subsec:phy-ara}

In addition to above two digital attacks (\ie, AQ-ARA and AP-ARA), we study a more important problem, \ie, whether the adversarially relighted face could be reproduced in the real world. 
%
\begin{figure}[t]
\includegraphics[width=\linewidth]{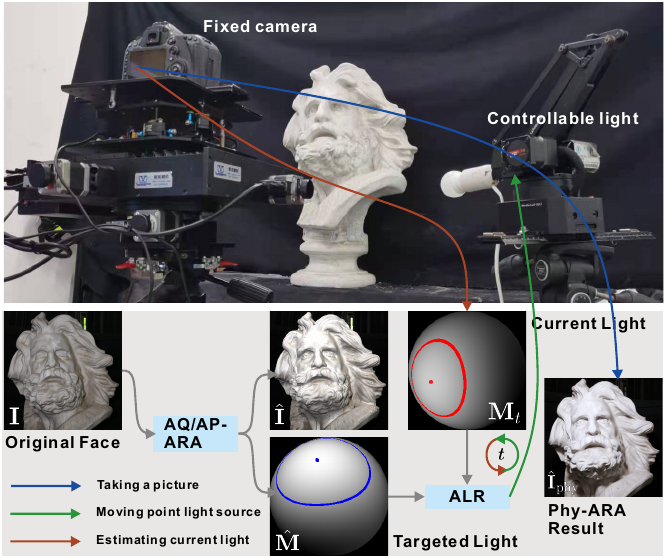}
\caption{Pipeline of Physical ARA. See text for details.}
\label{fig:phy}
\end{figure}
%
To this end, we propose the physical adversarial relighting attack (Phy-ARA). 
The main idea follows three steps: First, in a real-world scenario, we take a photo of a volunteer's face with a fixed camera under the natural light.
Then, we use the AQ-ARA or AP-ARA to perform the attacks under the guidance of a face recognition (FR) method and get an adversarially relighted face $\hat{\mathbf{I}}$ and the adversarial lighting $\hat{\mathbf{L}}$.
Finally, we can reproduce the adversarial lighting condition via a physical light source and take a new photo of the volunteer.

Although above process seems simple, one key issue makes it unavailable, that is, it is difficult to set suitable physical light sources meeting the pattern of estimated adversarial lighting.
We address this problem from two aspects. First, we use the commonly-used point light source (PLS) to simulate the estimated adversarial lighting. Second, following the state-of-the-art active lighting recurrence (ALR) method~\cite{our2018ALR}, we can physically adjust the position of the PLS by a robotic arm and produce the real lighting condition that is the same to the estimated adversarial one. \figref{fig:phy} shows the working scene and pipeline of Phy-ARA.

In contrast to existing ALR method \cite{our2018ALR} that depends on the parallel lighting model, our adversarial relighting is based on the spherical harmonics form that is a more general lighting representation and cannot be processed via the ALR directly.
To fill the gap, we change the generation manner of lighting map in the ALR and enable it to support our experiment. Specifically, given the estimated adversarial lighting $\hat{\mathbf{L}}$, we generate the respective lighting map by $\hat{\mathbf{M}} = \text{f}(\mathbf{N}^{\rm s},\hat{\mathbf{L}})$, where $\mathbf{N}^{\rm s}$ indicates the normal of a sphere. 
After that, given the $t$th frame of current camera, \ie, $\mathbf{I}_t$, the estimated scene normal $\mathbf{N}$ and the scene reflectance $\mathbf{R}$, we can calculate the corresponding spherical harmonics coefficients $\mathbf{L}_t$ by \reqref{eq:lamb}. Then, we calculate the lighting map of $\mathbf{I}_t$ by $\mathbf{M}_t = \text{f}(\mathbf{N}^{\rm s},\mathbf{L}_t)$ (See \figref{fig:phy}). 
The brightest position of $\mathbf{M}_t$ encodes the light source position in the azimuthal and polar axes, and the area of isointensity circle encodes the distance between light source and scene. Therefore, as shown in \figref{fig:phy}, we can get the instant navigation feedback from $\hat{\mathbf{M}}$ and $\mathbf{M}_t$. After that, we employ the incremental adjustment strategy in \cite{our2018ALR} to actively tune a robotic arm and update $\mathbf{L}_t$ to match $\hat{\mathbf{L}}$. Finally, we get the physical adversarial light (\ie, $\hat{\mathbf{L}}_\text{phy}$) under which we take a new image as the physical adversarial relighting image (\ie, $\hat{\mathbf{I}}_\text{phy}$). As shown in \figref{fig:phy}, we can clearly see that the image $\hat{\mathbf{I}}_\text{phy}$ basically has the same lighting distribution as the $\hat{\mathbf{I}}$, showing effectiveness of our physical experiment.

\begin{figure*}[t]
\centering
\includegraphics[width=\linewidth]{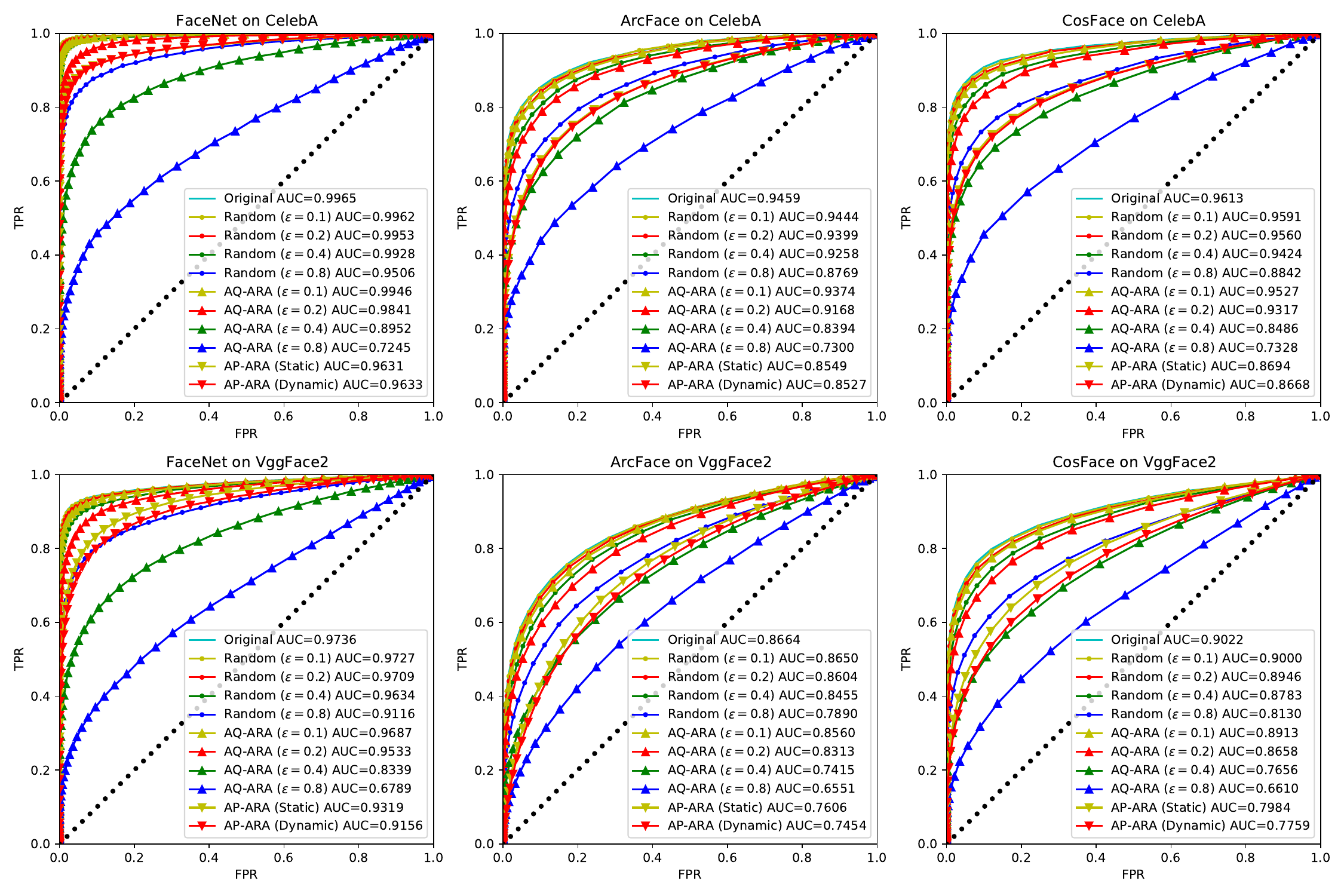}
\caption{ROC curves of various FR methods' performance under different attacks. The ``FaceNet on CelebA'' means that the adversarial examples generated by these attacks are tested by FaceNet method on CelebA dataset. ``AQ-ARA ($\epsilon=0.1$)'' indicates that the attack is launched with parameter $\epsilon=0.1$.}
\label{fig:roc}
\end{figure*}
%

\section{Experimental Results}\label{sec:exp}

\subsection{Setups}\label{subsec:setups}

{\bf Datasets.} 
Adversarial relighting aims to fool face recognition method and relates to two tasks, \ie, image relighting and face recognition. 
However, existing image relighting datasets do not contain the face identity annotations, thus are not suitable for evaluating our work. 
We conduct experiments on FR datasets, \ie, VGGFace2 \cite{cao2018vggface2} and CelebA \cite{liu2015deep}.
%
%
Both datasets can be employed to train ARNet of AP-ARA and evaluate the performance of FR attacks (See \secref{subsec:comp_attack} and \secref{subsec:comp_transfer}). 
In addition, the informative attribute annotations in the CelebA dataset support the statistical analysis of the adversarial examples generated by FR attacks (See \secref{subsec:comp_transfer}). 
For the attacking evaluation, we use test datasets of VGGFace2 and CelebA, including 169k images of 500 identities and 19,962 images of 1,000 identities, respectively. To ensure the validity of AP-ARA, we perform cross-validation between VGGFace2 and CelebA for the training and testing of AdvLNet.

{\bf Attack pipeline and metrics.} 
For an identity in the testing dataset, we have one of his/her face image $\mathbf{I}$ that may be took under arbitrary lighting sources. Then, given a face recognition model $\phi(\cdot)$, we can use the AQ-ARA (\ie, solving \reqref{eq:aqara} and \reqref{eq:lamb}) or AP-ARA (\ie, relying on a pre-trained network supervised by $\phi(\cdot)$) to relight $\mathbf{I}$ and produce a relighted face image $\hat{\mathbf{I}}$. 
After that, we can evaluate the attack performance to see whether a FR model can still achieve high performance on the relighted faces. When the evaluated FR model is the same with the model to guide or supervise AQ-ARA or AP-ARA, we mean it the white-box attack (See \secref{subsec:comp_attack}) otherwise we denote it as transfer-based attack (See \secref{subsec:comp_transfer}).

In terms of evaluation process, given a dataset for face recognition, we have $n$ identities and $2k$ face images per identity. For the $i$th identity, we separate the $2k$ face images into two subsets, \ie, the reference subset $\mathcal{R}_i$ and the targeted subset $\mathcal{T}_i$. Then, we use our ARAs and baselines to relight the face images in $\mathcal{T}_i$ and get adversarially relighted examples $\mathcal{A}_i$.
All the reference subsets $\mathcal{R}_i$ consist of a larger set $\mathcal{R}=\cup_{i=1}^n{\mathcal{R}_i}$ while all subsets $\mathcal{A}_i$ make up of another set $\mathcal{A}=\cup_{i=1}^n{\mathcal{A}_i}$
After that, we calculate the cosine similarity between face images from $\mathcal{R}$ and $\mathcal{A}$ based on a FR model. As a result, we can get a $nk\times nk$ matrix $\mathbf{S}$. 
A robust FR model is desired to have a $\mathbf{S}$ with the block-diagonal pattern, that is, the face images from the same identify should have high cosine similarity tending to $1$ otherwise having low similarities tending to $-1$.
We can also define a ground-truth matrix $\mathbf{G}\in \mathds{R}^{nk\times nk}$ for face recognition where $G(i,j)=1$ means the $i$th face in $\mathcal{R}$ and $j$th face in $\mathcal{A}$ are from the same identity otherwise $G(i,j)=0$.
We can calculate the true positive rate (TPR) and false positive rate (FPR) by comparing $\mathbf{S}$ with $\mathbf{G}$ under a given cosine similarity threshold. 
With a series of thresholds, we can draw the ROC curve and its area under curve (AUC) to measure the effectiveness of attacks, that is, a more effective adversarial relighting attack corresponds to a lower AUC.
In this paper, we select sixteen face images per identity, \ie, $k=8$.

In this paper, we use two blind image quality assessment metrics, \ie, BRISQUE \cite{Mittal2012} and NIQE \cite{mittal2012making}, to compare the image quality of generated adversarial examples. A larger BRISQUE or NIQE score indicates worse quality and less naturalness of an image.

{\bf Models.} We evaluate the proposed attacks and baselines against three face recognition models, \ie, FaceNet \cite{schroff2015facenet}, CosFace \cite{wang2018cosface}, and ArcFace \cite{deng2019arcface}. All these models require face images to be pre-processed by MTCNN \cite{zhang2016joint} for best performance, and we will do so as usual.

{\bf Baseline methods.} Our adversarial relighting only tunes the light parameters (\ie, nine spherical harmonics coefficients) to fool face recognition system, leading to smooth variation of the face image.
In contrast, existing additive-perturbation-based adversarial attacks can tune each pixel independently under the guidance of FR models \cite{fgsm,ifgsm,mifgsm}. 
For the fairness of the comparison, we do not include these attacks as part of the baselines.
Actually, a reasonable baseline is to conduct random relighting without any FR model guidance.
To this end, we can first randomly sample the nine lighting coefficient variations within the range $[-\epsilon,\epsilon]$ and use relighting methods to apply the sampled lights to the targeted face.
We employ deep portrait relighting (DPR) \cite{zhou2019deep} since it achieves the state-of-the-art relighting performance and show more realistic results than other methods, \eg, SfSNet \cite{sengupta2018cvpr}.
All experiments are conducted on a computer with an NVIDIA~RTX~2080 GPU.

\begin{table*}[t]
\centering
\begin{tabular}{c|c|ccc|ccc}
\toprule
\multicolumn{2}{c|}{Datasets} &
\multicolumn{3}{c|}{CelebA} &
\multicolumn{3}{c}{VGGFace2} \\
\midrule
\multicolumn{2}{c|}{Metrics} &
AUC & BRISQUE & NIQE &
AUC & BRISQUE & NIQE  \\
\midrule
\midrule
\multicolumn{2}{c|}{Original} &
0.9965 &
33.48 &
4.23 &
0.9736 &
41.70 &
5.01 \\
\midrule
\multirow{4}{*}{Random} &
$\epsilon=0.1$ &
0.9962 &
33.44 &
4.21 &
0.9727 &
41.59 &
4.99 \\
&
$\epsilon=0.2$ &
0.9953 &
33.40 &
4.21 &
0.9709 &
41.52 &
4.98 \\
&
$\epsilon=0.4$ &
0.9928 &
34.24 &
4.35 &
0.9634 &
41.74 &
5.04 \\
&
$\epsilon=0.8$ &
0.9506 &
38.10 &
4.92 &
0.9116 &
43.63 &
5.37 \\
\midrule
\multirow{4}{*}{AQ-ARA} &
$\epsilon=0.1$ &
0.9946 * &
33.24 &
4.17 &
0.9687 * &
41.32 &
4.94 \\
&
$\epsilon=0.2$ &
0.9841 * &
34.38 &
4.27 &
0.9533 * &
41.65 &
4.95 \\
&
$\epsilon=0.4$ &
0.8952 * &
39.87 &
4.83 &
0.8339 * &
45.23 &
5.35 \\
&
$\epsilon=0.8$ &
0.7245 * &
44.68 &
5.70 &
0.6789 * &
47.53 &
6.18 \\
\midrule
\multirow{2}{*}{AP-ARA} &
Static &
0.9631 * &
36.66 &
4.32 &
0.9319 * &
41.23 &
4.96 \\
&
Dynamic &
0.9633 * &
36.88 &
4.35 &
0.9156 * &
43.14 &
5.15 \\
\bottomrule
\end{tabular}
\caption{Performance and image quality results of baselines and our methods against FaceNet \cite{schroff2015facenet} on the CelebA and VGGFace2. We use * to mark the white-box attack.}
\label{tab:perf}
\end{table*}

\begin{figure*}[t]
\centering
\includegraphics[width=0.9\linewidth]{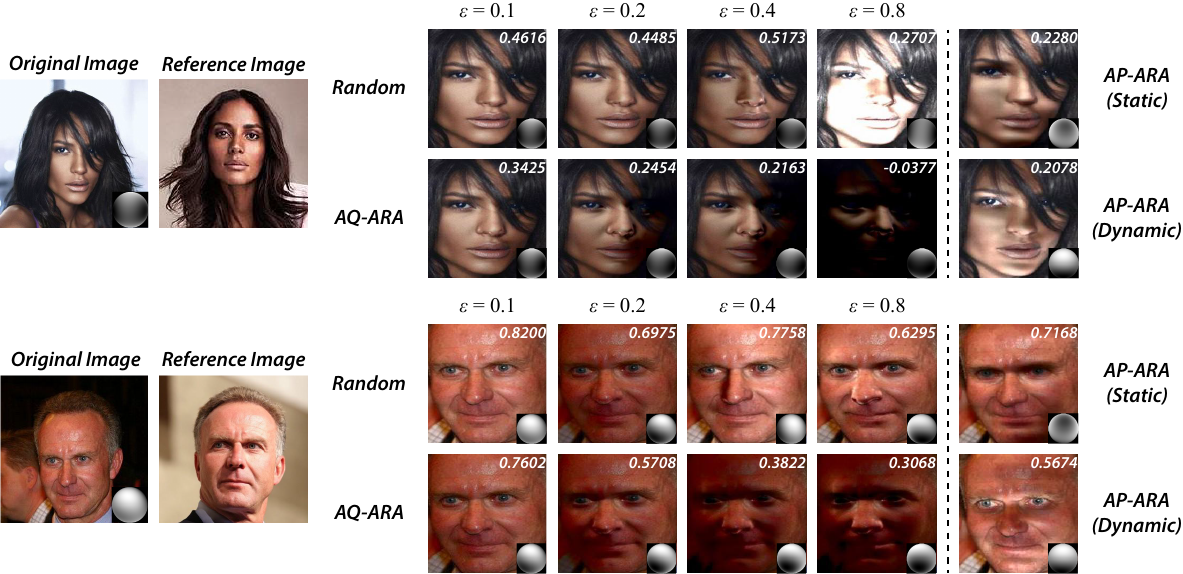}
\caption{Visualized adversarial examples of the proposed attacks and baselines. We vary the $\epsilon$ parameter to show the relighted faces and also place the shading sphere in the lower right corner of each face image.}
\label{fig:cases}
\end{figure*}

\subsection{Comparison Results on White-box Attack}
\label{subsec:comp_attack}

We conduct the proposed attacks and baselines against FaceNet on two datasets, \ie, CelebA and VGGFace2, and show the results in \tableref{tab:perf} and \figref{fig:roc}. For the random relighting baseline and the proposed AQ-ARA, we set the parameter  $\epsilon\in\{0.1,0.2,0.4,0.8\}$. The proposed AP-ARA is trained on CelebA and VGGFace2 datasets, respectively, and uses FaceNet as the FR method supervision, \ie, $\phi$ in \reqref{eq:lossadv}.
We have the following observations: \ding{182} As the norm ball (\ie, $\epsilon$) for the light becomes larger, the AUC under AQ-ARA gradually reduce with slight increasing of the BRISQUE and NIQE, demonstrating the effectiveness of the objective function \reqref{eq:aqara}.
\ding{183} When comparing the random relighting baseline with AQ-ARA, we see that AQ-ARA leads to lower AUC under the same $\epsilon$, demonstrating that AQ-ARA can is able to find adversarial light that misleads FaceNet easily . 
\ding{184} AP-ARA methods' AUCs are between the results of AQ-ARA with $\epsilon=0.2$ and $\epsilon=0.4$, but usually hold a better image quality (lower BRISQUE and NIQE) than AQ-ARA ($\epsilon=0.4$). In addition, the computational cost of AP-ARA is lower than that of AQ-ARA, due to the one-step ARNet in the test procedure.
\ding{185} When comparing the two variants of AP-ARA, AP-ARA-Dynamic gets lower AUC than AP-ARA-Static on VGGFace2 while having similar results on CelebA, hinting the effectiveness of the proposed ARNet that is able to adapt to different face via a dynamic convolution. 

We also provide the two visualization results in \figref{fig:cases} and have the following observations: 
\ding{182} Both AQ-ARA and AP-ARA are able to produce the realistic relighting pattern. More specifically, AQ-ARA tends to generate the adversarial light along a specific direction. For example, in second case, the adversarial light of AQ-ARA is along the northwest-southeast direction. The first case has similar result with a different direction. In contrast, AP-ARA tends to generate more complex light pattern that cannot be regarded as a directional light. 
\ding{183} When comparing the random relighting with AQ-ARA, we see that AQ-ARA always lets the FaceNet generate smaller cosine similarity under the same $\epsilon$, which further demonstrates the effectiveness of the proposed method. 
\ding{184} When comparing AP-ARA-Static with AP-ARA-Dynamic, the relighted faces of AP-ARA-Dynamic have lower similarity to the reference one. 

\begin{table*}[t]
\centering
\begin{tabular}{c|c|ccc|ccc}
\toprule
\multicolumn{2}{c|}{Datasets} &
\multicolumn{3}{c|}{CelebA} &
\multicolumn{3}{c}{VGGFace2} \\
\midrule
\multicolumn{2}{c|}{Transfer to} &
FaceNet & ArcFace & CosFace &
FaceNet & ArcFace & CosFace \\
\midrule
\midrule
\multicolumn{2}{c|}{Original} &
0.9965 &
0.9459 &
0.9613 &
0.9736 &
0.8664 &
0.9022 \\
\midrule
\multirow{4}{*}{Random} &
$\epsilon=0.1$ &
0.9962 &
0.9444 &
0.9591 &
0.9727 &
0.8650 &
0.9000 \\
&
$\epsilon=0.2$ &
0.9953 &
0.9399 &
0.9560 &
0.9709 &
0.8604 &
0.8946 \\
&
$\epsilon=0.4$ &
0.9928 &
0.9258 &
0.9424 &
0.9634 &
0.9455 &
0.8783 \\
&
$\epsilon=0.8$ &
0.9506 &
0.8769 &
0.8842 &
0.9116 &
0.7890 &
0.8130 \\
\midrule
\multirow{4}{*}{AQ-ARA} &
$\epsilon=0.1$ &
0.9946 * &
0.9374 &
0.9527 &
0.9687 * &
0.8560 &
0.8913 \\
&
$\epsilon=0.2$ &
0.9841 * &
0.9168 &
0.9317 &
0.9533 * &
0.8313 &
0.8658 \\
&
$\epsilon=0.4$ &
0.8952 * &
0.8394 &
0.8486 &
0.8339 * &
0.7415 &
0.7656 \\
&
$\epsilon=0.8$ &
0.7245 * &
0.7300 &
0.7328 &
0.6789 * &
0.6551 &
0.6610 \\
\midrule
\multirow{2}{*}{AP-ARA} &
Static &
0.9631 * &
0.8549 &
0.8694 &
0.9319 * &
0.7606 &
0.7984 \\
&
Dynamic &
0.9633 * &
0.8527 &
0.8668 &
0.9156 * &
0.7454 &
0.7759 \\
\bottomrule
\end{tabular}
\caption{Transferability of baselines and our methods against FaceNet \cite{schroff2015facenet}, ArcFace \cite{deng2019arcface}, and CosFace \cite{wang2018cosface} on the CelebA \cite{liu2015deep} and VGGFace2 \cite{cao2018vggface2}. All values in the table are the AUC scores on respective datasets. The adversarial examples are crafted from the FaceNet and we use * to mark the results of white-box attacks.}
\label{tab:tran}
\end{table*}

\subsection{Comparison Results on Transferability}
\label{subsec:comp_transfer}

We use the adversarial relighting examples crafted from FaceNet to attack other two face recognition methods, \ie, ArcFace \cite{deng2019arcface} and CosFace \cite{wang2018cosface}. 
As presented in the \tableref{tab:tran}, we see that: \ding{182} Both AQ-ARA and AP-ARA show significant transferability over the two datasets. Specifically, AQ-ARA and AP-ARA achieve similar even lower AUCs when attacking ArcFace and CosFace. 
\ding{183} The random relighting also presents some transferability. Nevertheless, the AUC reduction is much lower than the AQ-ARA under the same $\epsilon$, showing the advantages of our method for adversarial attack. 

%
\begin{figure*}[t]
\centering
\includegraphics[width=0.9\linewidth]{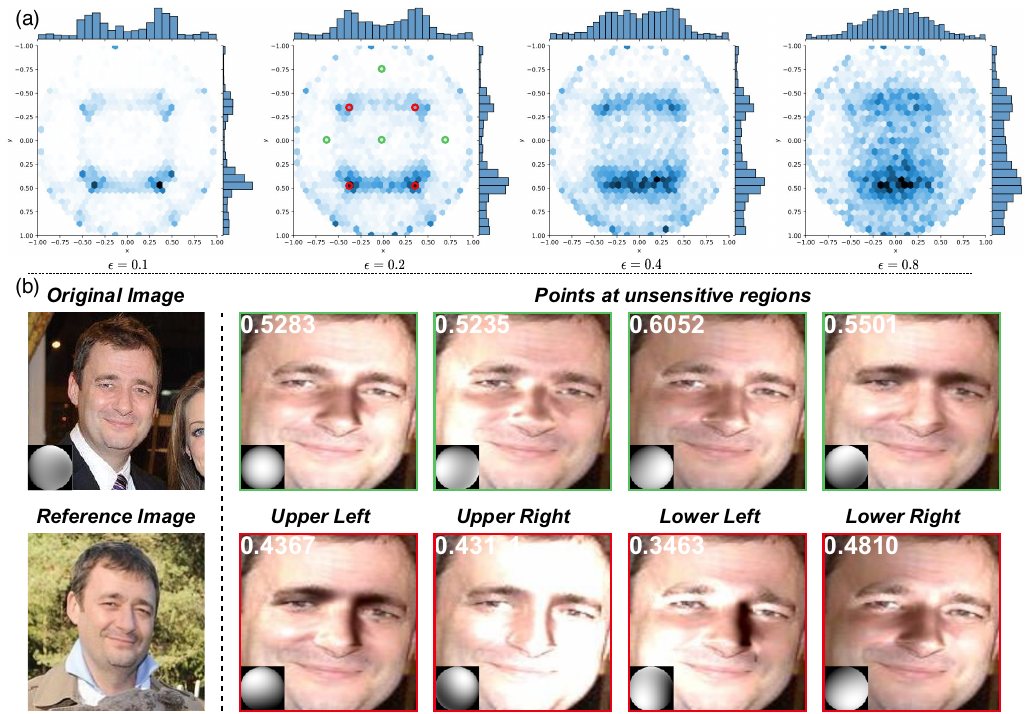}
\caption{(a) Four histogram maps under $\epsilon\in\{0.1,0.2,0.4,0.8\}$. Please find detailed descriptions in the text. (b) Relighting examples based on the light conditions defined in (a)-`$\epsilon=0.2$'. The similarity of all examples to the reference image are labeled at the upper-left corner.}
\label{fig:light_analysis}
\end{figure*}
%

\subsection{Analyzing effects of Light to Face Recognition via AQ-ARA}
\label{subsec:analysis}
In this section, we study the effects of light to FR via the proposed adversarial attack.
Specifically, for the $i$th example, we have a pair of adversarial light $\hat{\mathbf{L}}_i$ and the original light $\mathbf{L}_i$ via the AQ-ARA, and we denote their lighting maps as $\mathbf{M}_i$ and $\hat{\mathbf{M}}_i$, respectively. Then, we can calculate the difference between lighting maps of $\hat{\mathbf{L}}_i$ and ${\mathbf{L}}_i$ by $\mathbf{D}_i=|\hat{\mathbf{M}}_i-\mathbf{M}_i|$. 
and get the maximum difference position (\ie, sensitive points) on the map by $(x_i,y_i)= \argmax_{(x,y)}{\mathbf{D}_i}$.
After that, we can calculate a 3D histogram on hexagonal grids by counting the number of  sensitive points in each grid.
As a result, we can get four histogram maps under four $\epsilon$ and show them in \figref{fig:light_analysis}~(a). 
The higher values in the figure correspond to sensitive lighting sources that may fool face recognition easily.
We also provide relighting examples based on the sensitive and insensitive lighting conditions.

We have following observations: \ding{182} According to the difference maps under $\epsilon=0.1$ and $\epsilon=0.2$, the main differences locate at the four positions around the center, indicating the sensitive lighting sources to face recognition. For example, in \figref{fig:light_analysis}~(b), the sensitive lights indicated by red points lead to smaller similarity to the reference image.
\ding{183} As the $\epsilon$ becomes larger, the sensitive lighting sources increase because the larger $\epsilon$ allows larger variation of faces. Previous less sensitive lighting sources are also able to fool FR methods.
\ding{184} According to the results of $\epsilon=0.4$ and $\epsilon=0.8$, FR method is more sensitive to the lighting sources at the bottom.

\begin{figure*}[t]
\centering
\includegraphics[width=0.9\linewidth]{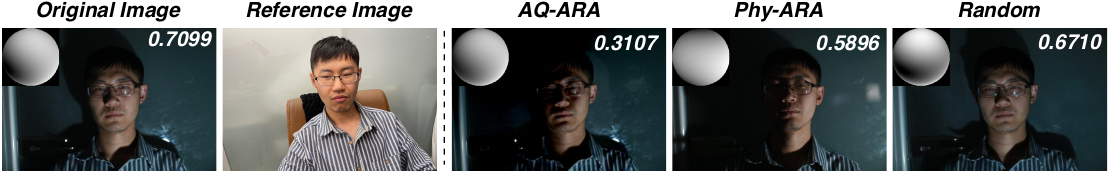}
\caption{Validation of Phy-ARA. See text for details.}
\label{fig:phy_res}
\end{figure*}
%

\subsection{Validating Physical Attack via Phy-ARA}

We follow the steps of \secref{subsec:phy-ara} to validate the proposed Phy-ARA through a volunteer.
Specifically, we first take an image of the volunteer with a natural light source as the original image.
Then, we conduct the AQ-ARA and produce a relighted face and the adversarial light.
After that, we drive a bulb fixed at a robotic arm to fit the adversarial light. 
Finally, we take a new picture as the result of Phy-ARA and calculate the cosine similarity between the reference image and other images based on the FaceNet.
As shown in \figref{fig:phy_res}, we see that: \ding{182} AQ-ARA's result is very similar to the physically relighted version, hinting that AQ-ARA is able to generate realistic relighted faces and the adversarial light is reproducible. \ding{183} Comparing the cosine similarity scores of the original and relighted images, we find that the adversarial light produced by AQ-ARA indeed affects FaceNet significantly, which reduces the similarity from 0.7099 to 0.3107. The respective physical counterpart reduces the similarity to 0.5896. In contrast, the random relighting affects the face recognition slightly.
%

\section{Conclusions}\label{sec:conc}

In this work, we have unveiled a new adversarial threat for FRS from the face lighting perspective and investigated a new task, the adversarial relighting attack (ARA). The proposed adversarial relighting aims at producing a high-realism relighted face image while being able fool the SOTA deep FR methods. We first designed the physical model-based ARA denoted as albedo-quotient-based adversarial relighting attack (AQ-ARA), which can generate natural adversarial light and synthesize adversarially relighted face images. To better suit efficiency-sensitive applications, we further proposed the auto-predictive adversarial relighting attack (AP-ARA) by training an adversarial relighting network to automatically predict the adversarial light in a one-step manner according to different input faces. More importantly, through a precise relighting device, we are able to transfer the above digital adversarial attacks to physical ARA (Phy-ARA), making the estimated adversarial lighting condition reproducible in the real world. The extensive evaluation of the proposed method on various SOTA FRS has demonstrated the feasibility of generating adversarially relighted faces to fool the FRS with ease. 
When the input faces are under other natural degradations such as very low-resolution, heavy occlusion, \etc, we expect that the proposed ARA would not be as effective. How to robustify the proposed ARA warrants a future study. Bad actors can potentially capitalize on the proposed ARA to fool some safety-critical FRS that are not yet prepared for this new type of attack. We hope that this work can also accelerate the R\&D of next-generation adversarially robust FRS. Moreover, we can use the proposed adversarial righting attack to analyze other face-related tasks, \eg, DeepFake detection \cite{qi2020deeprhythm}, visual-based heart rhythm estimation \cite{niu2020rhythmnet}, and facial age estimation \cite{feng2017feng}.






\ifCLASSOPTIONcaptionsoff
  \newpage
\fi

\bibliographystyle{IEEEtran}
\bibliography{ref}


%



\end{document}